# Lyapunov-based uncertainty-aware safe reinforcement learning

Ashkan B. Jeddi, Nariman L. Dehghani, and Abdollah Shafieezadeh

*Abstract*—**Reinforcement learning (RL) has shown a promising performance in learning optimal policies for a variety of sequential decision-making tasks. However, in many real-world RL problems, besides optimizing the main objectives, the agent is expected to satisfy a certain level of safety (e.g., avoiding collisions in autonomous driving). While RL problems are commonly formalized as Markov decision processes (MDPs), safety constraints are incorporated via constrained Markov decision processes (CMDPs). Although recent advances in safe RL have enabled learning safe policies in CMDPs, these safety requirements should be satisfied during both training and in the deployment process. Furthermore, it is shown that in memory-based and partially observable environments, these methods fail to maintain safety over unseen out-of-distribution observations. To address these limitations, we propose a Lyapunov-based uncertainty-aware safe RL model. The introduced model adopts a Lyapunov function that converts trajectory-based constraints to a set of local linear constraints. Furthermore, to ensure the safety of the agent in highly uncertain environments, an uncertainty quantification method is developed that enables identifying risk-averse actions through estimating the probability of constraint violations. Moreover, a Transformers model is integrated to provide the agent with memory to process long time horizons of information via the self-attention mechanism. The proposed model is evaluated in grid-world navigation tasks where safety is defined as avoiding static and dynamic obstacles in fully and partially observable environments. The results of these experiments show a significant improvement in the performance of the agent both in achieving optimality and satisfying safety constraints.**

## I. Introduction

Reinforcement learning (RL) has proven to offer great potential for a wide range of control tasks, such as video games (e.g., [1]), robot navigation (e.g., [2]), recommender systems (e.g., [3]), and resource management (e.g., [4]) among others. The underlying objective in these problems is to optimize the expected sum of rewards or costs through interaction of the agent with the environment. During this interaction, the agent has the freedom of exploration in order to find the optimal policy. However, in many real-world applications of RL, the granted freedom may lead to harmful actions without taking into consideration the safety-related consequences that threat both agent and the environment [5], [6]. Therefore, it is imperative to enforce the agent to avoid potentially harmful actions. An example that highlights safety concerns is the management of the power grid, where this critical infrastructure should not sustain any damage. In these safety-critical tasks, it is essential to find safe policies that not only optimize the expected sum of rewards or costs but also ensure the safe behavior of the agent during both training and deployment.

In order to guarantee the safety of the learned policy, constraints are imposed on the environment. This results in an extension of the Markov decision process (MDP) formulation which is used to describe the setup of RL problems to constrained Markov decision process (CMDP) formulation [7]. These constraints are often defined in terms of constraint-based cost functions separate from the reward-based objective function. In CMDPs with finite state and action spaces and when the transition model is unknown, the Lagrangian methods which are a classic approach for solving constrained optimization problems, are commonly applied (e.g., [8]–[10]). The underlying idea of the Lagrangian methods is to cast the CMDP problem into the unconstrained dual problem by penalizing the original objective function and solving the resulting function using primal-dual methods. Although convergence of the Lagrangian methods is guaranteed [7], no guaranty is provided for the convergence rate of these solutions. In addition, by implementing these methods, safety constraints are not usually satisfied during training [11]. Numerical instability of the primal-dual methods is another major drawback of the Lagrangian method which leads to constraint-violating behavior [12], [13]. Recently, Chow et al. [14], [15] used Lyapunov functions to explicitly model constraints in the CMDP framework and to develop a scalable safe RL approach.

Lyapunov functions have been ubiquitously used in the field of control (e.g., [16]) and optimization (e.g., [17]). A Lyapunov function assists in analyzing the stability (or asymptotic stability) of equilibrium points in a dynamical system. In the field of RL, Perkins and Barto [18] first proposed a Lyapunov-based RL method that applied state constraints on the action choices of the agent. Berkenkamp et al. [19] used Lyapunov functions to introduce the region of attraction (ROA) in model-based RL. ROA is a safe region in the state space where the

This work was supported by the National Science Foundation under Grant CMMI-2000156.

A. B. Jeddi is with the Department of Civil, Environmental, and Geodetic Engineering, Ohio State University, OH 43210 USA (e-mail: bagherijeddi.1@osu.edu).

N. L. Dehghani is with the Department of Civil, Environmental, and Geodetic Engineering, Ohio State University, OH 43210 USA (e-mail: laaldehghani.1@osu.edu).

A. Shafieezadeh is with the Department of Civil, Environmental, and Geodetic Engineering, Ohio State University, OH 43210 USA (e-mail: shafieezadeh.1@osu.edu).

equilibrium point is asymptotically stable. The underlying step in applying Lyapunov stability analysis to any control or optimization problem as well as RL problems is to find a Lyapunov function that satisfies the stability properties. This process is commonly implemented via deciding on a Lyapunov function candidate form and finding the parameters that hold the stability hypothesis [20]. However, one of the major shortcomings of this approach is that the design of Lyapunov functions is non-trivial. This shortcoming also applies to the case of RL problems and relevant studies have taken various approaches to arrive at a proper Lyapunov function [21]. In order to overcome this challenge in safe RL problems, Chow et al. [14], [15] proposed an LP-based algorithm to design Lyapunov functions for both discrete and continuous action spaces. Furthermore, using the Lyapunov functions designed by the proposed algorithm, Chow et al. [14] provided dynamic programming (DP) algorithms – namely, safe policy iteration (SPI) and safe value iteration (SVI) – for finite CMDP models. In addition, for the case of large state-space CMDP problems, Chow et al. [14] proposed safe deep $Q$-network (SDQN) which is an off-policy model-free fitted $Q$-iteration method. Although the proposed algorithm was shown to provide safe and optimal policies, it requires a feasible policy that already satisfies safety constraints to ensure the safety during training. In many complex real-world safe RL tasks, this feasible policy is not known a priori and therefore the application of SDQN algorithm is limited in these cases.

In RL problems, optimal policies can be learned either by using the collected data from prior real-world interactions [22] or by actively interacting with a simulated environment [4], [23]. In either case, RL algorithms have shown to respond poorly to real-world deployments and unseen data due to their failure to generalize over the training data [24], [25]. In addition, neural networks, which are generally used as function approximation tools either for *state value function* ($V$) or *state-action value function* ($Q$), act overconfident on out-of-distribution test data [6], [26]. This leads to the violation of safety constraints by the trained agents (i.e., resulting in unsafe policies). A group of studies in safe RL have focused on addressing these limitations via using statistical uncertainty information to take safe action decisions. In a prior study, Kahn et al. [27] used an uncertainty-dependent cost function to maintain safety in terms of avoiding collisions in unfamiliar environments. Lutjens et al. [28] further extended uncertainty-aware RL models by using Monte Carlo-dropout (MC-dropout) and bootstrapping in an ensemble of long-short-term-memory (LSTM) models to provide stochastic estimates of model uncertainty. Although LSTM models (generally recurrent architectures) add the ability of implicit memory, in practice, they show performance reductions in the case of long-term dependencies [29].

The issue of long-term dependencies is further highlighted in RL tasks with memory-based and partially observable environments. Autonomous driving is one of the prominent memory-based tasks where the agent should be capable of integrating information across frames to detect information such as velocity of objects [30]. To overcome the long-term dependency issues that LSTM models suffer from, Transformers architecture was introduced which process sequences entirely rather than sequentially. Transformers which were originally applied in the field of natural language processing and machine translation, removed the recursion process and added the ‘self-attention mechanism’ [29]. The introduced ‘attention’ value explicitly accounts for the importance of steps in a sequence and adds memory abilities. In the field of regular RL problems, while LSTM models have been adopted commonly (e.g., [1], [31], [32]), Transformer mechanism has been employed by only a few studies to provide memory for the agent. The application of Transformers model in those studies has shown superior performance compared to LSTM. In particular, Parisotto et al. [33] introduced the Gated Transformer-XL (GTrXL) by modifying the vanilla transformer and TransformerXL (TXL) [34] architecture. They further illustrated the superior performance of GTrXL in challenging memory-based tasks or partially observable environments compared to LSTM baselines. Recently, Parker et al. [35] adopted the adaptive attention span [36] idea to address the restricted memory block size in GTrXL by a variable length attention span. Although transformer based RL has shown promising performance in regular RL tasks, the use of transformer models and attention mechanism in satisfying long-term objectives while guaranteeing safety measures in safe RL domain has remained uncharted.

In this study, to address the problem of safety not only at convergence, but also throughout exploration and learning in both memory-based and reactive environments, we develop a new off-policy uncertainty-aware Transformers-based safe RL method. We use the formal formulation of safe RL problems in CMDP environments to define the problem of safe RL. In order to convert trajectory-based safety constraints to state-based constraints, we apply Lyapunov functions which provide theoretical safety guarantees. To overcome the need for a priori knowledge of a safe feasible policy and to ensure safety in memory intensive tasks during both training and testing, we further integrate the safe RL model with a Transformers-based encoder model with adaptive attention span and relative positional encoding. The introduced Transformers-based algorithm uses a sequence-to-sequence architecture that encodes information retrieved from the environment in terms of attention vectors and in turn adds memory abilities to the agent, while focusing on actions that are important in achieving reward-based objectives and/or satisfying the safety constraints. The encoded attention vectors are used in an ensemble of feed forward networks with MC-dropout and bootstrapping to provide qualitative predictive uncertainty estimates for the unseen data in terms of the probability of constraint violation. To ensure the safety of the actions, a risk-averse action selection scheme is introduced which selects the safest action using uncertainty estimates. We evaluate our algorithm in a set of fully and partially observable grid-world navigation tasks with static and dynamic obstacles in which the agent should avoid obstacles to satisfy the safety of the policy while minimizing its expected cumulative cost.

## II. Preliminaries

The objective of the developed safe RL approach is to provide policies that ensure the safety of the agent during both training and deployment in memory-based environments. To achieve this goal, we take advantage of the theoretical guarantees of Lyapunov-based algorithms for safety. Furthermore, to provide the agent with memory capabilities and be able to focus on important actions while accounting for long-term dependencies in the interaction with the environment, we employ a Transformers-based model. This model takes advantage of the self-attention mechanism. To provide an overview of the important components in our safe RL approach, in this section, after formalizing the problem of safe RL in CMDP models, an overview of the Lyapunov-based safe RL method is presented. Then, an overview of Transformers is provided, and finally, the required adjustments for employing Transformers in RL problems are discussed.

### A. Constrained Markov Decision Process

The problem of safe RL is formally formulated in a CMDP framework [7]. A CMDP is a tuple of $(X, A, P, c, d, x_0, d_0)$ where $X$ is the state-space, $A$ is the action-space, and $P(.|x,a)$ is the transition probability distribution. In this formulation, $c(x,a)$ is the immediate cost function and $d(x)$ is the immediate constraint cost. $d_0 \in \mathbb{R}_{\geq 0}$ is an upper-bound on the expected cumulative constraint cost and $x_0$ is defined as the initial state. We further define the cost function as $C_\pi(x_0) = \mathbb{E}\left[\sum_{t=0}^{T^*-1} c(x_t, a_t)|x_0, \pi\right]$ and the safety constraint function as $D_\pi(x_0) = \mathbb{E}\left[\sum_{t=0}^{T^*-1} d(x_t, a_t)|x_0, \pi\right]$. Given the safety constraint of the CMDP framework $D_\pi(x_0) \leq d_0$, we attempt to solve the following optimization problem:

$$\min_\pi \, [C_\pi(x_0)] : D_\pi(x_0) \leq d_0 \tag{1}$$

where the solution to this problem is denoted as the optimal policy, $\pi^*$. The Bellman optimality equation provides a recursive approach to find this optimal policy or the optimal value function. In this approach, by repeatedly applying a Bellman operator for the value function ($v$), a sequence of value functions is generated. According to the Banach fixed point theorem, as this Bellman operator is a contraction mapping, repeated application of the Bellman operator results in an optimal value function [37]. A generic version of the Bellman operator $\mathcal{B}_{\pi,h}: \mathbb{R}^{|X|} \to \mathbb{R}^{|X|}$ with respect to policy $\pi$ and arbitrary cost function $h(x,a)$ can be defined as:

$$\mathcal{B}_{\pi,h}[v](x) = \sum_a \pi(a|x) \left[ h(x,a) + \sum_{x'} P(x'|x,a) v(x') \right] \tag{2}$$

The Bellman operator is applied to the value function and in turn it produces a new value function. In a dynamic programming approach, this operator is applied repeatedly and eventually the process converges to the optimal value function using which an optimal policy can be derived. As this approach is computationally expensive, iterative methods such as value or policy iteration as well as temporal difference methods (e.g., $Q$-learning) are introduced. Furthermore, iterating over the Bellman operator does not necessarily lead to a policy that satisfies the safety constraint (i.e., $D_\pi(x_0) \leq d_0$). To provide solutions to this problem, methods such as safe RL using Lyapunov functions are developed.

### B. Safe Reinforcement Learning using Lyapunov Functions

Lyapunov functions are widely used in control theory to validate control strategies by showing the stability of the resulting behavior of the system. With the extension of this concept to the CMDP environment, the goal is to identify Lyapunov functions for the safe RL problem and then restrict the action choices of the agent so that the safety constraints are met. Based upon this principle, the off-policy model-free fitted $Q$-iteration SDQN method proposed by Chow et al. [14] entails designing a set of Lyapunov functions as follows:

$$\mathcal{L}_{\pi_B}(x_0, d_0) = \left\{ L: X \to \mathbb{R}_{\geq 0} : \mathcal{B}_{\pi_B,d}[L](x) \leq L(x), \forall x \in X; L(x_0) \leq d_0 \right\} \tag{3}$$

where $\pi_B$ is the feasible baseline policy. Given a function $L \in \mathcal{L}_{\pi_B}(x_0, d_0)$, the set of $L$-induced policies can be defined as:

$$\mathcal{F}_L(x) = \left\{ \pi(.|x) : \mathcal{B}_{\pi,d}[L](x) \leq L(x) \right\} \tag{4}$$

$\mathcal{F}_L(x)$ does not necessarily contain an optimal policy that satisfies Eq. (3). To address this limitation, Chow et al. [14] designed a Lyapunov function that guarantees the feasibility of the $L$-induced policies. This function can be designed as:

$$L_\epsilon(x) = \mathbb{E}\left[ \sum_{t=0}^{\infty} \gamma^t (d(x_t) + \epsilon(x_t)) | \pi_B, x \right] \tag{5}$$

where $\gamma$ is the discount factor and $\epsilon$ is an auxiliary constraint cost with the upper-bound of:

$$\epsilon^*(x) = \frac{2T^* D_{\max} D_{TV}(\pi^* || \pi_B)(x)}{1-\gamma} \tag{6}$$

where $D_{\max}$ is an upper bound for the immediate constraint cost and $T^*$ is a random variable representing the time of achieving the terminal state induced by policy $\pi^*$. $D_{TV}(\pi^*||\pi_B)$ indicates the total variation distance between the optimal policy, $\pi^*$, and feasible baseline policy, $\pi_B$, and is given by:

$$D_{TV}(\pi^*||\pi_B) = \frac{1}{2} \sum_a |\pi_B - \pi^*| \tag{7}$$

Using the bound $\epsilon^*$, a candidate Lyapunov function $L_{\epsilon^*}$ can be formed that satisfies Eq. (3). Without knowing $\pi^*$ a priori, to implement a scalable method of the developed safe RL approach, $\epsilon^*$ is approximated with an auxiliary constraint cost $\tilde{\epsilon}$. To ensure the inclusion of the optimal policy in $\mathcal{F}_L$, the approximated auxiliary constraint cost should be optimized to find the largest value that satisfies $\mathcal{B}_{\pi_B,d}[L_{\tilde{\epsilon}}](x) \leq L_{\tilde{\epsilon}}(x)$ and the safety condition $L_{\tilde{\epsilon}}(x_0) \leq d_0$. Therefore, this value needs to be calculated via solving the following LP problem:

$$\tilde{\epsilon} \in \operatorname{argmax} \left\{ \sum_{x \in X} \epsilon(x) : d_0 - D_{\pi_B}(x_0) \geq \mathbf{1}(x_0)^\top \left( I - P(x'|x, \pi_B) \right)^{-1} \mathbf{1}(x) \epsilon \right\} \tag{8}$$

where $\mathbf{1}(x)$ is a one-hot vector with non-zero values at $x$. Furthermore, the defined Lyapunov function in Eq. (5) is constructed using the concept of state-action value via a function approximation. Leveraging the function approximation is essential for the case of large or continuous state-space where accurate definition of immediate constraint cost function $d(x_t)$ is non-trivial. Deep $Q$-networks are used for approximating Eq. (5) as follows:

$$\hat{Q}_L(x, a: \theta_D, \theta_T) = \hat{Q}_D(x, a: \theta_D) + \tilde{\epsilon}. \hat{Q}_T(x, a: \theta_T) \tag{9}$$

where $\hat{Q}_D$ and $\hat{Q}_T$ are the constraint and stopping time value networks. Inclusion of the $\hat{Q}_T$ values enforces the agent to select a policy that safely achieves the terminal state with fewer number of actions. Moreover, using function approximation and the baseline policy $\pi_k$, the auxiliary constraint cost which is the solution to Eq. (8) can be estimated as:

$$\tilde{\epsilon} = \frac{d_0 - \pi_k(.\,|x_0)^\top \hat{Q}_D(x_0, .: \theta_D)}{\pi_k(.\,|x_0)^\top \hat{Q}_T(x_0, .: \theta_T)} \tag{10}$$

Using the formulated state-action Lyapunov function $\hat{Q}_L$ and the auxiliary constraint cost $\tilde{\epsilon}$, the action probabilities $(\pi')$ can be calculated by the following LP problem:

$$\pi'(.\,|x) \in \text{argmin}_\pi \left\{ \pi(.\,|x)^\top Q(x,a) : \left(\pi(.\,|x) - \pi_B(.\,|x)\right)^\top \hat{Q}_L(x,a) \le \tilde{\epsilon} \right\} \tag{11}$$

At each step of the $Q$-iteration, the updated policy parameter $(\phi^*)$, which defines the optimal safe policy $(\pi^*)$, can be calculated through policy distillation (Rusu et al., 2015) from the calculated action probabilities $(\pi')$ as:

$$\phi^* \in \text{argmin}_\phi \frac{1}{m} \sum_{m=1}^{M} \sum_{t=0}^{T^*-1} D_{\text{JSD}} \left( \pi_\phi(.\,|x_{t,m}) || \pi'(.\,|x_{t,m}) \right) \tag{12}$$

where $M$ is the size of the batch of state trajectories generated by the baseline policy $\pi_B$. The adopted policy distillation method creates a mapping between the optimal action probabilities which are the solution of Eq. (11) and the updated policy $\pi^*(x|a: \phi^*)$. Moreover, $D_{\text{JSD}}$ is the Jensen-Shannon divergence. A more detailed discussion on the SDQN algorithm can be found in the study by Chow et al. ([14]).

### C. Transformers Architecture and Variants

The issue of long-term dependencies in learning sequential data has highlighted the inefficiency of recurrent models (e.g., LSTM) when there is a large gap in time between the relevant information and point of query. Unlike recurrent models, Transformers process sequential data entirely and they effectively account for the long-term dependencies by the introduction of attention mechanism. The Transformer model has an encoder-decoder structure. The encoder component provides an encoding of the input sequence. This process is implemented via the use of several stacked blocks of encoders with a layer of multi-head attention module followed by a position-wise multi-layer feed forward network with weights $\boldsymbol{W}_o$. The multi-head attention layer is a refinement of the self-attention layer which adds $h$ representation subspaces working in parallel. In this context, self-attention is defined as:

$$\text{Attention}(Q, K, V) = \text{softmax}\left(\frac{QK^\top}{\sqrt{d_k}}\right) V \tag{13}$$

where $Q$, $K$, and $V$ denote the query vector, key vector, value vector, which are calculated via learning the weights $\boldsymbol{W}_q$, $\boldsymbol{W}_k$, and $\boldsymbol{W}_v$, respectively. Furthermore, $d_k$ is the dimension of key vectors. Through this mechanism, similarities of token $t$ with past token $r$ in span $[t-S, t)$ with length $S$ is calculated as:

$$s_{tr} = x_t^\top \boldsymbol{W}_q^\top (\boldsymbol{W}_k x_r + p_{t-r}) \tag{14}$$

where $p_{t-r}$ is the relative positional embedding. Then, attention weights are obtained as a function of these similarities via applying a softmax function as:

$$a_{tr} = \frac{\exp(s_{tr})}{\sum_{q=t-S}^{t-1} \exp(s_{tq})} \tag{15}$$

The outputs of attention heads are then calculated by using the attention weights $a_{tr}$ and value weights $\boldsymbol{W}_v$ as:

$$y_t = \sum_{r=t-S}^{t-1} a_{tr} \boldsymbol{W}_v x_r \tag{16}$$

The input to the self-attention layer is an embedding of the input sequence. In the case of language modeling, this step provides an embedding of the word while for the RL problem, this is an embedding of the observations in time steps of the rollout. The multi-head attention module applies the self-attention mechanism and takes advantage of residual connections [38] and layer normalization [39]. Moreover, the feed forward network applies a temporal 1 by 1 convolution. Residual update and layer normalization are applied to the outputs of this feed forward network as well.

One of the major shortcomings of the vanilla Transformer, especially the multi-head attention layer, is its permutation invariance which is resolved to some extent via the use of positional encodings [29]. However, this solution only provides limited context-dependency due to limiting the maximum dependency distance to the fixed length of the input. Furthermore, due to the context fragmentation where every segment is trained separately, no dependency is captured between these segments. To overcome these limitations, Dai et el. [34] introduced Transformer-XL which improved upon the vanilla Transformer by extending the context size compared to what was previously possible. This was achieved by applying a recurrent memory scheme similar to a hidden state in a recurrent neural network (RNN) model. The recurrence mechanism, which adds the memory scheme to the model, involves inputting the output of the previous hidden layer for the previous segment in addition to the output of the previous hidden layer for the current segment which was originally used in the vanilla Transformer structure. The disadvantage of this recurrence model is a change in the positional encoding of the input tokens, while same positional encoding is expected from the tokens of different segments. To overcome this confusion, Transformer-XL introduced the relative positional encoding that is a part of each self-attention layer and uses the relative distance between tokens instead of their absolute distance.

Although the Transformer-XL model has been able to achieve promising performance, it suffers from computational inefficiency during both training and inference. With the aim of a more computationally efficient Transformer model, Sukhbaatar et al. [36] introduced the concept of adaptive attention span. This concept involves learning the attention span from the data directly. For this purpose, the integer value of the attention span is converted to a continuous value through a soft-masking function. The soft-masking ($m_z$) is defined as a function of the masking length ($z$), which is applied to each attention head, and is learned from the data during back-propagation. This model was shown to be able to retain a larger memory and learn long-term dependencies all without increasing computation cost. The adaptive attention span is of great importance, particularly in RL problems, as the rollout sequence length of the interactions with environments varies in separate experiments. Moreover, as the actions taken by the agent in later steps of individual experiments may be more significant for the reward-based objective, it is imperative to adopt an adaptive attention span in the Transformers model for the case of RL problems. Due to the soft-masking function (i.e., $m_z$), the attention weights defined in Eq. (15) are modified as:

$$a_{tr} = \frac{m_z(t-r)\exp(s_{tr})}{\sum_{q=t-S}^{t-1} m_z(t-q)\exp(s_{tq})} \tag{17}$$

The performance of Transformer models is highly dependent on the size of the feed forward module. With a focus on simplifying the architecture of Transformer models, Sukhbaatar et al. [40] introduced the all-attention layer which merges the attention and feed forward layers into a single unified attention layer. The results from this model showed its capability in achieving a similar performance to Transformer-XL on language modeling tasks with about half the number of parameters.

### D. *Transformers in Reinforcement Learning*

Despite the performance of Transformer model and its variants in natural language processing (NLP) tasks, large body of literature in RL problems has not taken advantage of the full potential of these models. In one of the early studies, Mishra et al. [41] showed that both the vanilla Transformer and Transformer-XL structure cannot achieve a better performance than a random action selection. With the aim of alleviating this random performance and stabilizing the performance of Transformer models in RL tasks, Parisotto et al. [33] introduced GTrXL with applying architectural modifications to the encoder block of the Transformer-XL model. One of these modifications is identity map reordering which is an alteration of the layer normalization order. Identity map reordering includes placing of the layer normalization before the multi-head attention and feed forward network, which results in a smoother optimization in the Transformer block. Furthermore, Parisotto et al. [33] introduced gating layers. A wide variety of gating mechanisms were applied in this model, and it was shown that a GRU-type gating scheme [42] produces the highest performance. In this scheme, reset gate is defined as:

$$r = \sigma(W_r y + U_r x) \tag{18}$$

where $x$ and $y$ are the input from skip connections and output from the multi-head attention module. $\sigma$ is the sigmoid activation function. Furthermore, the weights of the reset gate and other gating layers which are denoted by $W$ and $U$ are not tied in depth. The update gate is computed as:

$$z = \sigma\left(W_z y + U_z x - b_g\right) \tag{19}$$

where $b_g$ is the bias in the gate. The candidate activation $\hat{h}$ is then calculated as:

$$\hat{h} = \tanh\left(W_g y + U_g(r \odot x)\right) \tag{20}$$

where $\odot$ is an element-wise multiplication. The activation of the GRU is a linear interpolation between the input from skip connections and the candidate activation $\hat{h}$ in the form of:

$$g = (1-z) \odot x + z \odot \hat{h} \tag{21}$$

The mentioned modifications enabled the GTrXL model to achieve state of the art performance and large improvement gains in both MDP and POMDP environments in multi-task DMLab30 which consists of both reactive and memory-based environments [33]. In this study we use GTrXL with GRU-type gating in adjunction to the all-attention scheme which takes advantage of the adaptive attention span and persistent memory.

## III. Methodology

In this section, we present the Lyapunov-based uncertainty-aware safe RL algorithm that uses a Transformers model to retain a memory of the environment and account for long-term dependencies in a rollout sequence. Moreover, uncertainty information is used in a safe action selection algorithm to avoid constraint violations during both training and test time in uncertain environments and for new observations. The proposed system architecture is depicted in Fig. 1. In this model, the agent observes the state of the environments (entirely or partially) along with other information from the environments (i.e., cost and constraint cost). An encoder model takes the returned information from the environments as an input and in return provides a memory mechanism with the use of multi-head attention layers and all-attention model. Then, the output of this encoder model (i.e., the encoded attention vectors) is used as an input to the constraint violation prediction model which is an ensemble of feed forward networks with bootstrapping and MC-dropout. The constraint violation prediction model provides predictions of the probability of constraint violation. Sample mean and variance of these predictions for each action are used to select the safest action. This objective is achieved through introduction of a risk-averse action selection mechanism according to the estimate of the probability of constraint violation and the uncertainty around it. Then, the agent uses this safe action to interact with the environment. In parallel, the encoded attention vectors are used in the Lyapunov-based SDQN algorithm. In return, the off-policy SDQN model provides the optimal safe policy.

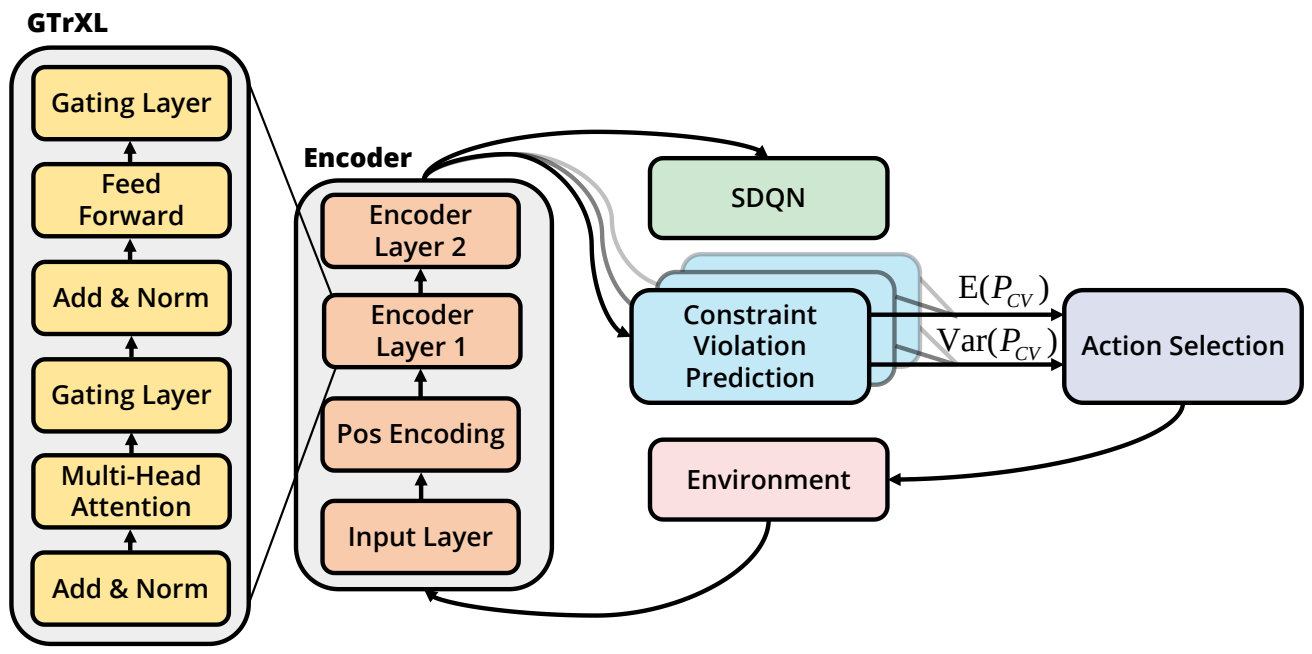

Fig. 1. The proposed model architecture for the uncertainty-aware safe RL

### A. Encoder Model

The encoder model is composed of an input layer, a relative positional encoding layer, and a stack of two identical GTrXL encoder layers. The input layer maps the input sequence from the environment (discrete or image observations) to a vector with hidden dimension $d_h$ through a fully-connected feed forward network. Relative positional encoding, which is inspired by Vaswani et al. [29], encodes the sequential information from the input layer through sine and cosine functions by element-wise addition of the relative positional encoding and the input vector. The output of the relative positional encoding layer is fed into two GTrXL encoder layers. Each encoder layer consists of residual connections, layer normalizations, multi-head attention, gating layers, and position-wise feed forward networks.

### B. Constraint Violation Prediction Network

The constraint violation prediction model takes as input the encoded attention vector $g(\boldsymbol{x}_{0:t}, \boldsymbol{a}_{0:t}, \boldsymbol{c}_{0:t}, \boldsymbol{d}_{0:t})$ and outputs the probability of violating the specified constraint $P_{CV}$ in the horizon $h$. By using the encoded attention vector, we account for the long-term dependencies in the trajectory and focus on the important actions that lead to constraint violation. Therefore, in comparison with the case of the direct utilization of the information obtained from the interaction with environment (e.g., [27]), more accurate estimates of the probability of constraint violation are expected using the proposed approach. The input of this model consists of attention values for the state $\boldsymbol{x}_{0:t}$, actions $\boldsymbol{a}_{0:t}$, cost $\boldsymbol{c}_{0:t}$, and constraint costs $\boldsymbol{d}_{0:t}$ of the rollout sequence and the output is defined as:

$$P_{CV} = P(\mathbf{1}_{CV} = 1 | \boldsymbol{x}_{0:t}, \boldsymbol{a}_{0:t}, \boldsymbol{c}_{0:t}, \boldsymbol{d}_{0:t}, \boldsymbol{a}_{t:t+h}) \quad (22)$$

where $\mathbf{1}_{CV}$ is a constraint violation label and $\boldsymbol{a}_{t:t+h}$ is the evaluated actions from the current time step $t$ to horizon $h$. The predicted probability $P_{CV}$ values are only estimates of the true probability of constraint violation due to the uncertainties in both static and dynamic environments (especially in the case of partially observable settings and new observations). Therefore, the effects of these uncertainties in environment should be considered by uncertainty estimates of $P_{CV}$. In order to obtain accurate uncertainty estimates, two techniques of Bootstrapping and MC-dropout are used here.

Bootstrapping provides uncertainty estimates of the data via resampling. Given a dataset $H$, $B$ new datasets $(H_b)$ are sampled with replacement such that the size of the selected sample is equal to the original dataset $H$. Then, an ensemble consisting of $B$ models $(NN_b)$ is used on each individual sampled dataset and the output of each model is an uncertain prediction of the true output that would have been acquired using a single model on the entire dataset. The expected prediction and uncertainty estimates are calculated as the sample mean and variance of the outputs from the ensemble of models. It is worth mentioning that the number of bootstrapped models used in uncertainty estimation has a large impact on the overall computational cost of the proposed model. To alleviate this computational cost, bootstrapping is used in conjunction with dropout. Dropout is a common method for reducing overfitting effects during training. In this method, at each forward pass through the network, the activation value of each hidden unit is set to 0 with the probability of $p$. The applied Bernoulli mask during training therefore prevents the units in network $NN_b$ from co-adapting on the training data by constructing a new randomized version of network, $NN_{b_d}$. On the other hand, at test time, the applied dropout regularization is removed and instead all the weights of hidden units learned during training are normalized by $p$ to compensate for the overpredicted weights. Gal and Ghahremani [43] showed that dropout can be used as an effective and computationally cheap method of acquiring a variational lower bound of uncertainty estimates if applied at test time.

### C. Risk-averse Uncertainty-aware Action Selection

Uncertainties in environment in RL settings may lead to the selection of actions that violate the safety of both the agent and environment. Therefore, it is essential to incorporate uncertainty-awareness measures that ensure safe exploration rather than randomly sampled exploration actions. However, only a handful of studies have been able to develop uncertainty-aware safe action selection approaches. For example, Kahn et al. [27] proposed a safe RL algorithm through introducing an uncertainty-aware safe exploration model to define a constraint-based cost. This cost, multiplied by a penalty coefficient, is then integrated with the RL objective as a penalization term. While the proposed safe RL model aims to drive safe exploration via selection of the safest action, it has been shown that penalization of the RL objective through a fixed penalty method (Lagrange multiplier) induces numerical instability in learning safe policies, which can lead to harmful actions [11], [44]. Lutjens et al. [28] addressed this limitation by introducing a safe action selection module independent of the RL objective. However, their proposed action selection module is limited to fully observable environments where the exact value of first-hitting time of the terminal state (i.e., $T^*$) is observed. In this study, we extend the action selection method proposed by Lutjens et al. [28] to partially observable environments. This is achieved by defining an action selection module independent of the cost associated with the time to reach the goal and adopting the concept of stopping time value network, $\hat{Q}_T$, to provide estimates of $T^*$ in the SDQN module. Furthermore, the risk-averse uncertainty-aware action selection module in this study enables the experience replay buffer to include a larger set of possible action sequences, which consequently leads to a more optimal policy selection. The action selection module introduced in this study selects the safest risk-averse action for the horizon $h$ with the minimum uncertainty-aware joint cost as:

$$a^*_{t:t+h} = \text{argmin}_{a\in A}\left(\lambda_e \mathbb{E}(P_{CV}) + \lambda_v \sqrt{\text{Var}(P_{CV})}\right) \quad (23)$$

where $\lambda_e$ and $\lambda_v$ are the weights of the first and second order moment cost terms, respectively. These values determine the relative importance of uncertainty in the selection of the safe actions and are treated as hyperparameters. In a highly uncertainty-averse action selection model, where $\lambda_v$ is set to a large value compared to $\lambda_e$, explorative actions are discouraged. This in turn prohibits learning the optimal policy. In this study, as the safety of the agent throughout the exploration process is guaranteed via the SDQN module, a relatively small value is defined for the weight of the second order moment cost terms (i.e., $\lambda_v = -2000$). Therefore, the agent is capable of returning higher values of reward during early phases of training.

Using the modules in this section, the overall algorithm of the proposed framework is presented in Algorithm 1.

**Algorithm 1** Lyapunov-based Uncertainty-aware Safe RL

**Input:** Initial experience replay buffer $M = \{\emptyset\}$; Initial importance weights $w_0 = 1$, $w_{D,0} = 1$, $w_{T,0} = 1$; Mini-batch size $|B|$; Network parameters $\theta^-$, $\theta^-_D$, and $\theta^-_T$
**for** $k \in \{0,1,\ldots\}$ **do**
  **for** $t = 0$ **to** $T^* - 1$ **do**
    Select *risk-averse action* $a^*_{t:t+h}$ execute and observe next state, $x_{t+1}$ and costs $(c_t, d_t)$
    *GTrXL*: calculate encoded attention $g(\boldsymbol{x}_{0:t}, \boldsymbol{a}_{0:t}, \boldsymbol{c}_{0:t}, \boldsymbol{d}_{0:t})$
    Add experiences to replay buffer $M \leftarrow g(x_{0:t}, a_{0:t}, c_{0:t}, d_{0:t}) \cup M$
    Sample a mini-batch $B = \{g(x_j, a_j, c_j, d_j, x'_j, w_j, w_{D,j}, w_{T,j})\}$ for $j = 1, \ldots, |B|$
    Set targets $y_{D,j}$, $y_{T,j}$, and $y_j$ as:
    $y_{D,j} = d(x_j) + \pi_k(\cdot\,|x'_j)^{\top} \hat{Q}_D(x_j, \cdot; \theta_D)$
    $y_{T,j} = \mathbf{1}\{x_j \in \mathcal{X}'\} + \pi_k(\cdot\,|x'_j)^{\top} \hat{Q}_T(x_j, \cdot; \theta_T)$
    $y_j = c(x_j, a_j) + \pi'(\cdot\,|x'_j)^{\top} \hat{Q}(x_j, \cdot; \theta)$
    where $\pi'(\cdot\,|x'_j)$ is the greedy action probability w.r.t. $x'_j$
    Update:
    $\theta_D \leftarrow \theta^-_D - \alpha_j w_{D,j}\left(y_{D,j} - \hat{Q}_D(x_j, a_j; \theta^-_D)\right)\nabla_\theta \hat{Q}_D(x_j, a_j; \theta)$
    $\theta_T \leftarrow \theta^-_T - \alpha_j w_{T,j}\left(y_{T,j} - \hat{Q}_T(x_j, a_j; \theta^-_T)\right)\nabla_\theta \hat{Q}_T(x_j, a_j; \theta)$
    $\theta \leftarrow \theta^- - \beta_j w_j \left(y_j - \hat{Q}(x_j, a_j; \theta^-)\right)\nabla_\theta \hat{Q}(x_j, a_j; \theta)$
    Update importance weight of the samples in mini-batch, based on TD errors
    Update the policy to $\pi_{k+1}$ using Policy distillation w.r.t. data $\{x'_{0,j}, \ldots, x'_{T^*-1,j}\}$ and $\{\pi'(\cdot\,|x'_{0,j}), \ldots, \pi'(\cdot\,|x'_{T^*-1,j})\}$ for $j = 1, \ldots, |B|$
  **end for**
  Update $\theta^- = \theta$, $\theta^-_D = \theta_D$, and $\theta^-_T = \theta_T$ after $c$ iterations
**end for**

## IV. Experiments

This section investigates the performance of the proposed model in a series of stochastic 2D grid-world navigation tasks with static and dynamic obstacles based on the DeepMind AI safety grid-worlds [45]. In these environments, the objective of the agent is to travel safely to a specific destination. At each time step, the agent can take the action to move to one of its four adjacent states or do not move and stay in the current state. However, to introduce uncertainty in the environment stemming from sensing and control noise, the agent will move to a random adjacent state with the probability of $\delta$. In both of the static and dynamic obstacle environments, an error probability of $\delta = 0.05$ is used. In between the starting point and the destination, a number of obstacles are placed. The agent can pass through but should avoid the obstacles for safety. For every time the agent hits an obstacle, a constraint cost of 1 is incurred. In these navigation tasks, the safety constraint is defined as hitting the obstacles by at most $d_0$ times in a pass from the starting point to the destination. The reward of reaching the destination is set to be equal to 1000. Furthermore, to encourage the agent to choose the shortest possible path, a stage-wise cost of 1 is defined for the actions of the agent.

### A. *Grid-world with Static Obstacles*

In this section, we evaluate the performance of the model in a grid-world with static obstacles. Two types of possible input observations are used in the case of this environment, namely discrete and image observation. The discrete observation is a one-hot encoding of the position of the agent in the grid. The image observation is a snapshot that shows entire RGB 2D grid map. For discrete observations, we use a feed forward neural network with hidden layers of size 16, 64, 32, and ReLU activations. For image observations, we use a convolutional neural network with filters of size $3 \times 3 \times 3 \times 32$, $32 \times 3 \times 3 \times 64$, and $64 \times 3 \times 3 \times 128$, with $2 \times 2$ max-pooling, each followed by ReLU activations. We then pass the result through a 2-hidden layer network with sizes 512 and 128. The Adam optimizer with the learning rate of 0.0001 is applied. At each iteration, we collect an episode of experience (100 steps) and perform 10 training steps on batches of size 128 which are sampled uniformly from the replay buffer. We update the target Q-networks every 10 iterations and the baseline policy every 50 iterations.

In the 2D grid-world environment with static obstacles, an obstacle density ratio $\rho \in (0,1)$ is defined that determines the number of obstacles with respect to the size of the grid-world. The navigation task is therefore more challenging if $\rho$ is closer to 1. Fig. 2 presents the results of return and cumulative constraint cost in terms of the obstacle density for the proposed framework along with SDQN and two Lagrangian-based approaches i.e., PPO-Lagrangian and TRPO-Lagrangian. In the Lagrangian-based approaches, the Lagrange multiplier is optimized via extensive grid search. Image observations are used for these tests and a constraint threshold of $d_0 = 5$ is selected. The presented empirical results show the superior performance of the proposed model in terms of the optimality and safety compared to the other safe RL approaches. The TRPO-Lagrangian method is incapable of achieving any feasible policy even for smaller values of obstacle density. On the other hand, the PPO-Lagrangian is capable of satisfying the safety constraint for almost all obstacle densities, while it achieves significantly lower return values compared to the proposed model and SDQN. By comparing the performance of the proposed model and the SDQN model, it can be concluded that the proposed safe RL method can achieve a more optimal policy while satisfying the safety constraints. This in mainly attributed to the ability of memory provided by the attention mechanism in the proposed safe RL method. The advantage of attention mechanism becomes even more evident in the case of more challenging environments ($\rho > 0.3$) where the agent has been able to achieve higher return values by focusing on the actions that are important in reaching the destination. Moreover, as the introduced action selection model determines the safest action through the use of probability of constraint violations, the constraint violation cost is significantly lower than other benchmark models over all density ratios.

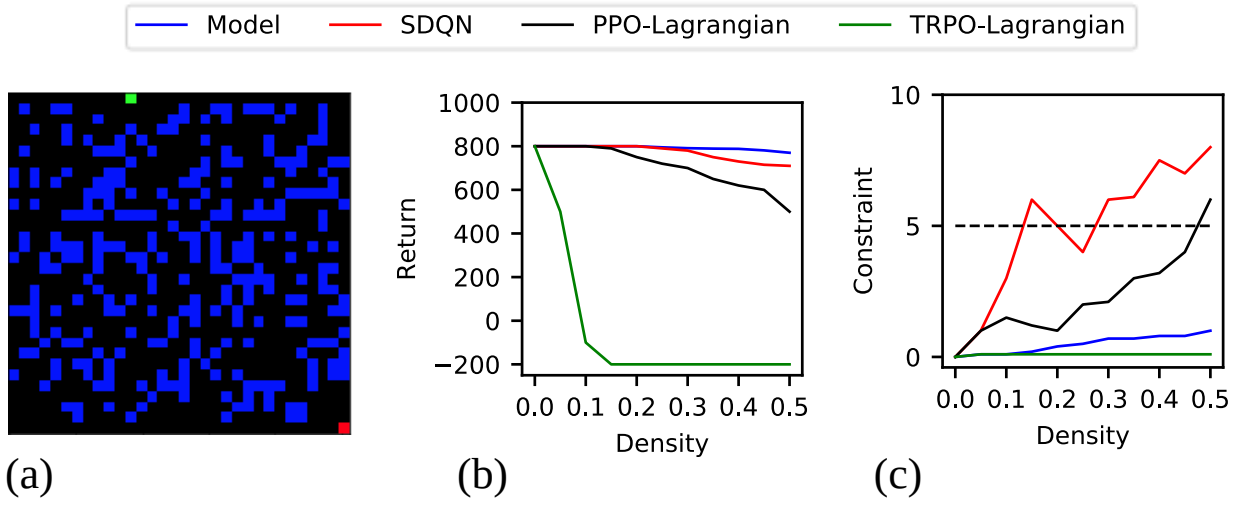

Fig. 2. Results of the grid-world navigation with static obstacles during training. (a) 2D grid-world environment ($\rho = 0.3$), (b) average return, and (c) average cumulative constraint cost.

To further evaluate the performance of the proposed model in a variety of the environments with different safety constraints, we compare the model with $d_0 = 1$ and $d_0 = 5$ for both discrete and image observations. We again compare the performance of the model with SDQN, TRPO-Lagrangian and PPO-Lagrangian methods. The results of this comparison are provided in Fig. 3 in terms of rewards and constraint cost during training. Results show that the studied Lagrangian-based methods are incapable of achieving a safe policy and the numerical instability of these methods has resulted in taking unsafe actions during training. In general, these methods often achieve lower rewards compared to SDQN and the proposed model, and they are more inclined to violate the constraints during training. These results further show that the SDQN model is capable of achieving a safe policy only at the end of training, while the proposed model can satisfy the safety constraints even in the beginning episodes of training. This is critical in satisfying safety as violating constraints even in a few episodes of training can be harmful to the agent and environment. Another interesting observation is that the proposed model is capable of reaching to higher rewards after a few episodes in the training and can achieve the maximum possible reward value earlier than the other methods. Moreover, it can be seen that the proposed model can achieve a better performance in the case of image observations compared to discrete observations. This trend exists because a snapshot of the entire grid map encapsulates relatively more information compared to a one-hot encoding of the position of the agent, which can be used in the Transformers model to have a rich memory of the environment. In addition, the case of $d_0 = 5$ offers more freedom to the agent, thus, the model can achieve the highest possible reward earlier than the more challenging case of $d_0 = 1$.

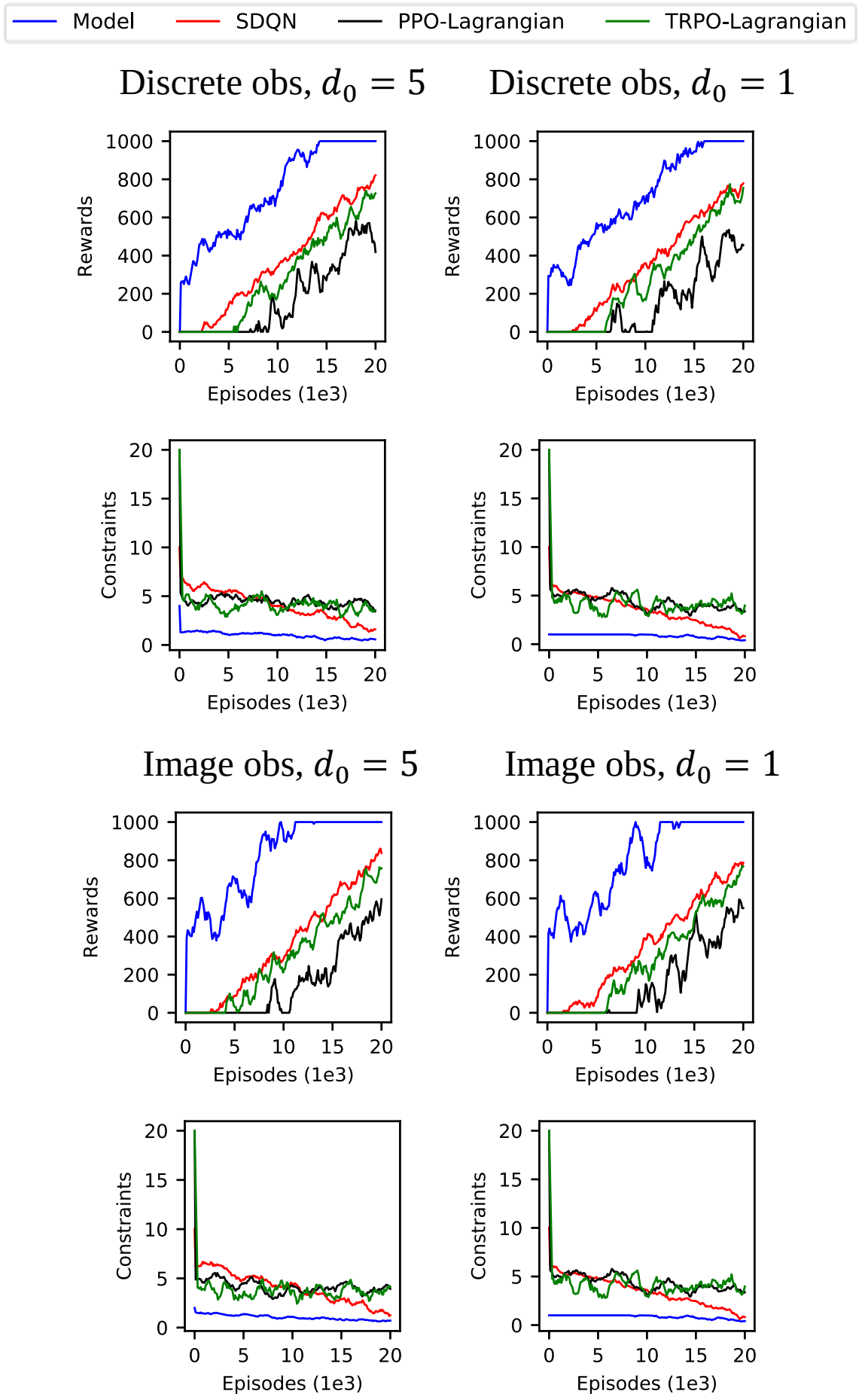

Fig. 3. Results of the grid-world navigation for discrete and image observations with static obstacles

## B. *Grid-world with Dynamic Obstacles*

In this section we evaluate the performance of the proposed model in a partially observable environment. The setting of this environment is illustrated in Fig. 4(a) for a $16 \times 16$ grid map. The objective of this agent is to travel from the starting point to the specified destination. There are eight moving dynamic obstacles in this grid-world which randomly move to one of their four neighboring states at each time step. The agent is able to observe the grid and the obstacles in a moving window of $5 \times 8$. The safety constraint in this environment is defined as hitting the dynamic obstacles less than $d_0 = 10$ times. The results of training of the agent using the proposed model compared to other safe RL methods are illustrated in Fig. 4 in terms of the reward and constraints. According to Fig. 4, the proposed method is able to achieve a safe and optimal policy. This is mainly attributed to the capability of the model to form a memory of the environment due to the implemented Transformers. While other Lagrangian-based models have been unable to achieve a safe policy at all, the SDQN model has determined a safe policy at the end of training. However, this implies that the SDQN approach will violate safety constraints during training.

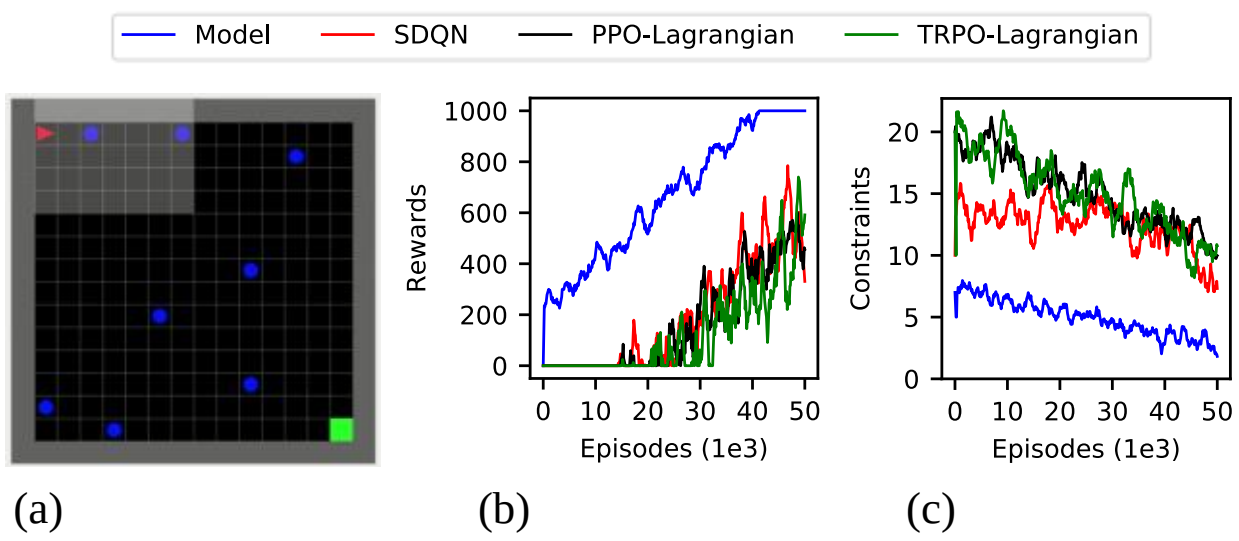

Fig. 4. Results of the grid-world navigation for partial image observations with dynamic obstacles. (a) 2D grid-world environment, (b) rewards, and (c) constraint violations

In order to gain a better insight toward the multi-head attention values, a visualization of these values is presented in Fig. 5. In these results, attention weights that are assigned to the actions in the last time step of the trajectory in the attention span are illustrated. Furthermore, the behavior illustrated here is captured after 10,000 episodes; therefore, the agent has formed a memory of the environment at this time step. According to this figure which provides a relative comparison of attention weights associated with different actions in global directions, the agent focuses mostly on important actions. For example, in Fig. 5(a), the Transformers model assigns the largest attention weights to the action 'right', 'down', and 'stay'. Action 'right' in this state leads to an immediate constraint cost and action 'down' is the action which moves the agent toward the destination. While the partial observations that are provided to the agent do not contain the exact position of the destination, the memory provided by the Transformers model guides the agent toward the destination. In this time step, the agent also assigns some attention weights to the action 'stay' which implies that staying at the same state might be important in reaching the destination or avoiding an immediate constraint cost. On the other hand, the attention weights indicate that action 'stay' is not important in the state depicted in Fig. 5(b). Although the added memory capability is effective in the determination of important actions, the information explicitly provided by the observation overwrites the attention weights. This can be seen in Fig. 5(c) where the exact position of the specified destination is observable and hence the attention weights indicate action 'down' as the most important action which leads to high reward values.

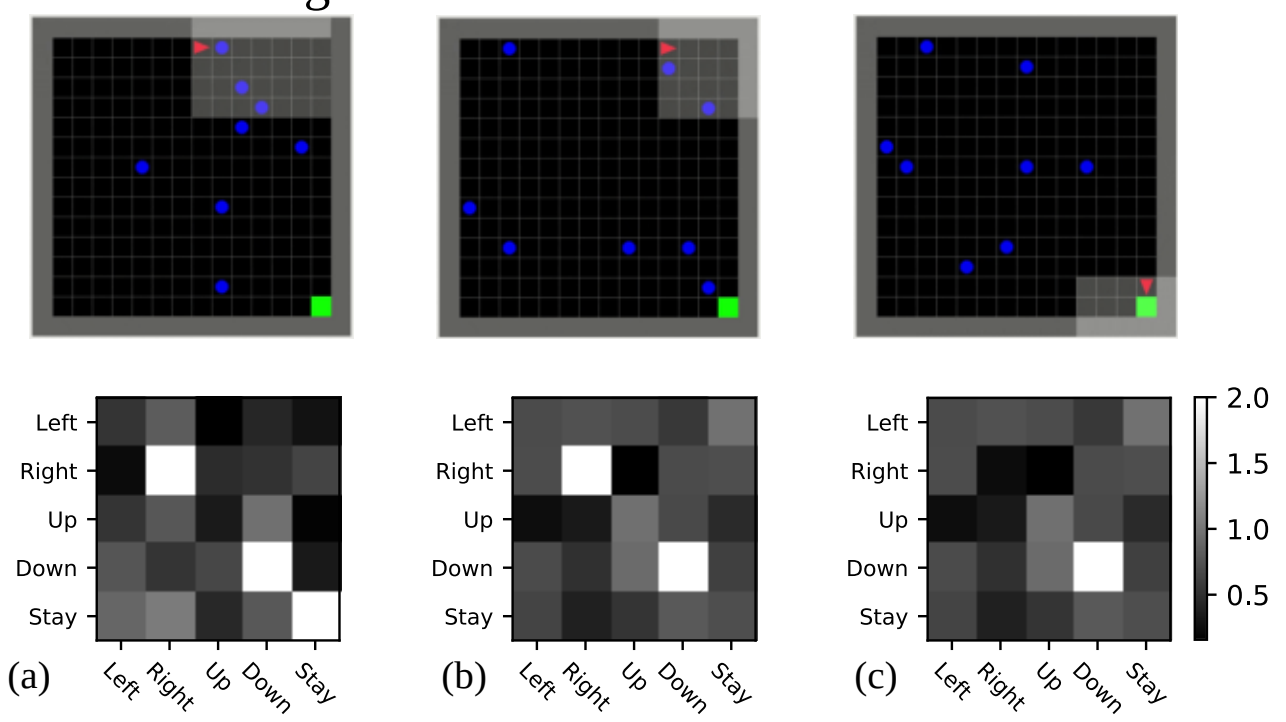

Fig. 5. Visualization of the multi-head attention weights in grid-world with dynamic obstacles at different states

## V. Conclusion

In this study, we introduced a novel uncertainty-aware safe reinforcement learning (RL) algorithm through the use of Lyapunov functions and Transformer models. We used the constrained Markov decision process (CMDP) environment to formally define the safe RL problem and define safety constraints. A Lyapunov-based policy updating safe RL approach was used to convert trajectory-based to state-based constraints and achieve the optimal safe policy. In order to account for the uncertainty of the environment and new observations during test time, we used a Transformers-based approach to provide uncertainty estimates of the environment. Moreover, the inclusion of Transformers model has added the ability of memory to the agent which was shown to be useful especially in the case of partially observable environments. The uncertainty estimates were achieved through the use of MC-dropout and bootstrapping in an ensemble of feed forward neural networks. These uncertainty estimates were obtained in terms of the probability of constraint violation. An action selection approach was introduced based on the uncertainty estimates in order to select risk-averse safe actions in the interaction with the environment.

We evaluated the performance of the proposed model in a series of navigation tasks in 2D grid-world environments. The objective in these tasks was to navigate from the starting point to the specified destination. During this travel, the agent should avoid hitting obstacles and therefore the safety in this task was introduced in terms of obstacle avoidance. These grid-world environments had static and dynamic obstacles. In the case of dynamic obstacles, we limited the observation of the agent to a bounded bow around the agent to impose partially observable settings. Through comparisons with state-of-the-art safe RL methods (i.e., SDQN, Lagrangian-PPO, Lagrangian-TRPO), we showed the superior performance of the proposed approach in terms of reward and constraint violations (i.e., a more optimal and safer policy) during training and at deployment. Furthermore, we showed that the inclusion of the Transformers-based encoder module provides a memory for the agent, which results in a performance improvement in partially observable environments.

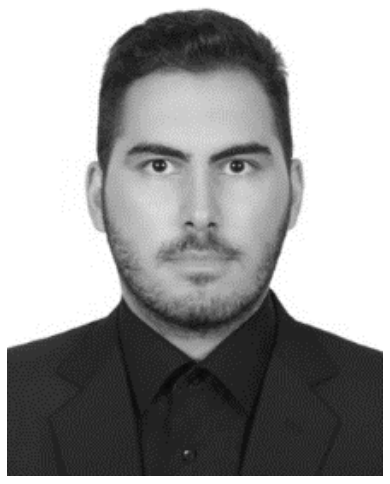

**Ashkan B. Jeddi** received the B.S. degree in civil engineering from Iran University of Science and Technology, Tehran, Iran, in 2015., the M.S. in structural engineering from Sharif University of Technology, Tehran, Iran, in 2017. He is a Graduate Research Associate in the Department of Civil, Environmental and Geodetic Engineering, The Ohio State University, Columbus, OH, USA. His primary research interests include application of machine learning techniques for the risk and resilience assessment of structural and infrastructure systems.

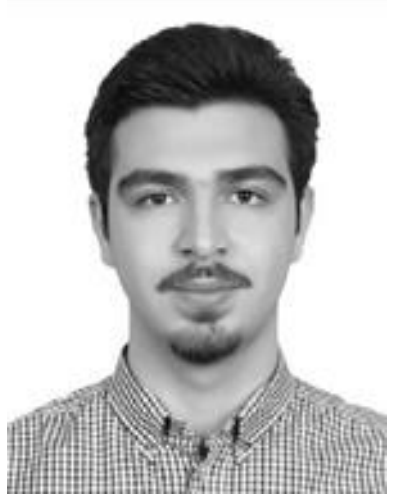

**Nariman L. Dehghani** received the B.S. degree in civil engineering from Iran University of Science and Technology, Tehran, Iran, in 2013. He received the M.S. degree in earthquake engineering from Amirkabir University of Technology, Tehran, Iran, in 2016. He is currently pursuing the Ph.D. degree in structural engineering at The Ohio State University, OH, USA. He is a Graduate Research Associate in the Department of Civil, Environmental and Geodetic Engineering, The Ohio State University, Columbus, OH, USA. His primary research interests include sustainability and resilience quantification, reliability assessment of civil infrastructure systems, and optimal decision making for multi-state systems.

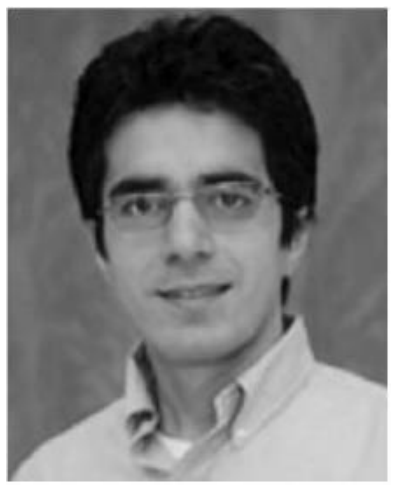

**Abdollah Shafieezadeh** received the B.S. and M.S. degrees in civil engineering from the University of Tehran, Tehran, Iran, in 2002 and 2006, respectively. He received the M.S. degree in structural engineering from the Utah State University, Logan, UT, USA, in 2008 and the Ph.D. degree in structural engineering with a minor in mathematics from the Georgia Institute of Technology, Atlanta, GA, USA, in 2011.

He is an Associate Professor in the Department of Civil, Environmental and Geodetic Engineering, The Ohio State University, Columbus, OH, USA. His primary research and teaching interests are in resilient systems, uncertainty quantification, and decision making in uncertain environments. He is a Sam Nunn Security Fellow awarded by the Sam Nunn School of International Affairs at Georgia Institute of Technology.